\pdfoutput=1

\documentclass[11pt]{article}

\usepackage[]{acl}

\usepackage{times}
\usepackage{latexsym}
\usepackage{siunitx}

\usepackage[T1]{fontenc}

\usepackage[utf8]{inputenc}

\usepackage{microtype}

\usepackage{graphicx}
\usepackage{amsmath}
\usepackage{breqn}
\usepackage{makecell}
\usepackage{tabularx}
\usepackage{booktabs}
\usepackage{multirow}
\usepackage[colorinlistoftodos]{todonotes}
\usepackage{textcomp}

\usepackage{bm}

\usepackage{hyperref}
\usepackage{amsmath}
\newcommand{\ForumCounts}{
\begin{table}[t!]
\small
\centering
\begin{tabular}{c c | cc} \Xhline{2\arrayrulewidth}
             \textbf{Forum} & \textbf{Frequency}& \textbf{Forum} & \textbf{Frequency} \\ \Xhline{\arrayrulewidth} 
             
            English &	662	&	Travel	&	356	\\
            Cooking	&	636	&	Music	&	262	\\
            Gaming	&	485	&	Bicycles	&	242	\\
            SciFi	&	408	&	DIY	&	213	\\
            ELL	    &	378	&	Aviation & 190	\\
              
             \Xhline{2\arrayrulewidth}
            
\end{tabular}
\caption{The ten most frequent forums found in the AnswerSumm dataset and their associated counts.}
\label{tab:num_sources}
\end{table}
}

\newcommand{\IntroductionMotivatingExample}{
\begin{table}[h!]
\centering
\small
\begin{tabularx}{\columnwidth}{|X|}
\hline
\textbf{Question:} I recently relocated to USA and have no Credit Score. Is Secure Credit Card is the only option for me to start building my credit score? Also please recommend which other credit cards are available for people like me to build credit score \\ \hline
\textbf{Answer 1:}  If you have an \textcolor{blue}{AMEX from another country, you can get an AMEX in the US. American Express has a separate system} that is not as strongly country-dependent as, say, VISA and MasterCard... \\ \hline
\textbf{Answer 2:} Secured credit cards are usually not very cost effective for building credit. \textcolor{red}{Find a local credit union}, of medium to large size. A credit union is like a bank, but operates under slightly different rules, and is non-profit...  \\ \hline
\textbf{Answer 3:} If you have had an \textcolor{blue}{American Express card abroad, you can try and get a US Amex}... \\ \hline
\textbf{Answer 4:} If the country you came from has \textcolor{green}{an HSBC}, you can ask \textcolor{green}{HSBC to use your credit rating from that country to give you an HSBC Mastercard} in the US... \\ \hline
\textbf{Summary:} \\ \hline
There are a range of options available to you, although your chance of success will depend on the bank that you apply with. However, if you have previously had a card with \textcolor{green}{HSBC} or \textcolor{blue}{American Express}, the process may be simpler. Other options could include \textcolor{red}{borowing from a credit union} or asking a friend or family member to be an additional cardholder with you. \\
\hline
\end{tabularx}
\caption{An example summary from our AnswerSumm dataset, illustrating the multiple viewpoints present manually-written summaries, and a subset of the 8 user answers to which the summary can be aligned.}
\label{tab:example}
\end{table}
}

\newcommand{\DatasetComparison}{
 \begin{table}[t]
\resizebox{\columnwidth}{!}{\begin{tabular}{c c c c c}
\Xhline{2\arrayrulewidth}
             \textbf{Dataset} & \textbf{Novel unigrams} &  \textbf{Ext. Oracle}  &  \textbf{Input Len}  & \textbf{Summ Len}    \\  \hline
            \textbf{AnswerSumm} & 21.0 &  40.05/18.45/35.70 & 787 & 47  \\ 
            XSUM & 35.8 & 29.79/8.81/22.65 &  431 & 23 \\
            CNN & 16.8 & 50.38/28.55/46.58 & 761 & 46 \\
            DailyMail & 17.0 & 55.23/30.55/51.24 & 653 & 55 \\
            \Xhline{2\arrayrulewidth}
\end{tabular}}
\caption{Comparison between AnwerSumm and the XSum \cite{narayan-etal-2018-dont} and CNN-DailyMail \cite{nallapati-etal-2016-abstractive} datasets, with data statistics from \cite{narayan-etal-2018-dont}. Oracle Extractive and Length refer to the maximum ROUGE \cite{lin-2004-rouge} score achievable by an extractive model, and the average length of the input and summaries, respectively.}
\label{tab:statistics}
\end{table}
}

\newcommand{\DatasetStatisticsAnswerSumm}{
 \begin{table}[t]
\centering
\resizebox{0.8\columnwidth}{!}{\begin{tabular}{c | c c}
\Xhline{2\arrayrulewidth}
           \textbf{Task} &  \textbf{Input}  &  \textbf{Output}   \\ \hline
           \multirow{3}{*}{SentSelect}  & 6.4 Ans & \multirow{3}{*}{9.2 Sents} \\ 
            & 40.3 Sents &  \\
            & 787 Words &  \\ \hline
           SentCluster & 9.2 Sents & 2.6 Clusters \\ \hline
            \multirow{2}{*}{ClusterSumm} & 3.4 Sents & \multirow{2}{*}{21 Words}  \\ 
            & 77 Words &  \\ \hline
           ClusterSummFusion & 55 Words & 47 Words \\ \hline
            \Xhline{2\arrayrulewidth} 
\end{tabular}}
\caption{Average statistics for input and output across the four Answersumm subtasks. E2ESumm's input is that of the SentSelect and the output is that of ClusterSummFusion.}
\label{tab:statistics_answersumm}
\end{table}
}

\newcommand{\ManualPipelineFigure}{
\begin{figure*}[t]
    \centering
    \includegraphics[width=\textwidth,height=5\textheight,keepaspectratio]{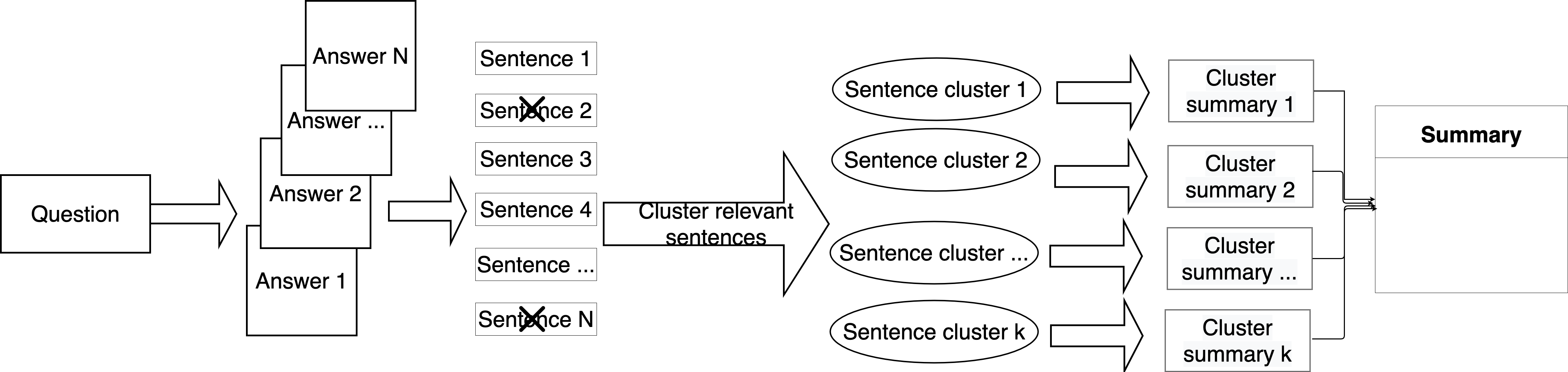}
    \caption{An illustration of our dataset annotation pipeline. Given a question and answers to that question, professional linguists 1) select relevant sentences, 2) cluster those selected sentences, 3) summarize each cluster's sentences, and 4) fuse clusters into a coherent, overall summary.}
    \label{fig:manual_pipeline}
\end{figure*} 
}
\newcommand{\PipelineFigure}{
\begin{figure*}[t]
    \centering
    \includegraphics[width=2.0\columnwidth,height=5\textheight,keepaspectratio]{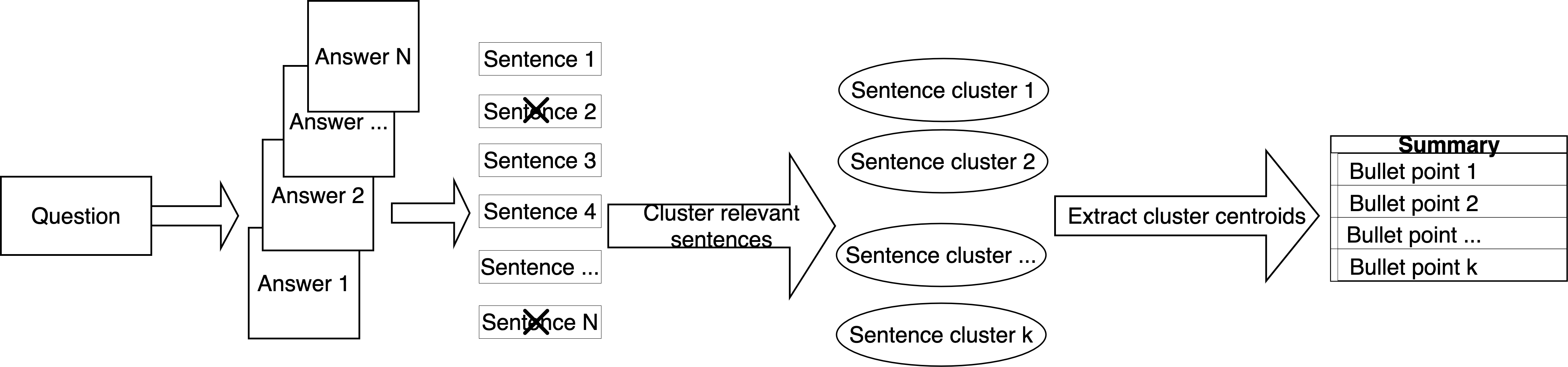}
    \caption{An illustration of our automatic dataset pipeline which mirrors the manual pipeline for data augmentation. Given a question and answers, relevant sentences are selected and clustered. Then, the cluster centroid sentence of non-singleton clusters is removed from the input to use as bullet point summaries.}
    \label{fig:pipeline}
\end{figure*} 
}

\newcommand{\ExampleSummaries}{
\begin{table}[h!]
\small
\centering
\begin{tabularx}{\columnwidth}{|X|}
\hline
\textbf{Question:} I bought a house that had been sitting for some time.  One of the the issues that I've discovered is that the flapper valve leaks.  I've come to this conclusion by turning off the water to the tank and combing back later to see that the water level is lower.  I have already replaced the flapper valve itself, and the leak remains.
 \\ \hline
\textbf{Answer 1:} Sometimes when a flapper gets old and begins to fail (disintegrate) it can leave a piece behind stuck to the outflow pipe that it covers. This piece/remnant then prevents the new flapper from getting a good seal. You might want to clean out the tank and make sure there are no remnants of the old flapper stuck in there. 
\\ \hline
\textbf{Answer 2:}
Over time the surface of the plastic part that joins the tank to the bowl can get tiny defects in it that      prevent the flapper from making a good seal. As Jeff suggests, you could try cleaning that part, or just replace it.
\\ \hline
\textbf{Answer 3:} I recently had two toilets begin to leak and had a hard time figuring out exactly where. I hate plumbing but decided to do full replacement of the various parts. I purchased two toilet repair kits for about \$18-20 each. Instructions on the package explained what to do. When you are done all parts and gaskets that wear or deteriorate over time are replaced and you essentially have a new toilet... \\ \hline
\textbf{Answer 4:} You might consider replacing most of the insides of your toilet... \\ \hline
\hline
\textbf{BART Summary:} The options are to replace the flapper, clean out the tank and replace it with a new piece, replace the entire tank or replace the whole tank with Fluidmaster's S2DBL or Home Depot's toilet flapper.
 \\ \hline
\textbf{BART-aug Summary:} There are a number of options available, including replacing the flapper, cleaning out the tank, replacing the line from the shut-off to the tank or replacing the entire toilet.
 \\ \hline
\textbf{BART-aug+RL Summary:} It is possible that the old flapper is stuck in the outflow pipe and is causing the leak. You could try cleaning the tank to remove any remnants. Alternatively, Home Depot sells a new flapper that comes with a new p     iece that it rests on.
 \\ \hline
\end{tabularx}
\caption{Sample input and model outputs. The outputs are factually consistent and cover multiple perspectives present in the input answers.}
\label{tab:example_summaries_1}
\end{table}
}

\newcommand{\RelResults}{
\begin{table}[t!]
\small
\resizebox{\columnwidth}{!}{\begin{tabular}{|c|c|c|}
\hline
    & \textbf{True Rel}   & \textbf{True Not Rel}   \\ \hline
Predicted Rel	& 4324	& 3349  	\\ \hline
Predicted Not Rel	& 5664 & 25088	\\ \hline
\end{tabular}}
\caption{RoBERTa confusion matrix on SentSelect.}
\label{tab:rel_results}
\end{table}
}

\newcommand{\ClusterSumm}{
\begin{table}[t!]
\small
\centering
\resizebox{.8\columnwidth}{!}{\begin{tabular}{|l|c|}
\hline
    \textbf{Task}   & \textbf{ROUGE-1/2/L}   \\ \hline
ClusterSumm	&	30.98/10.61/26.22	\\ \hline
ClusterSummFusion	&	51.64/32.67/47.13	\\ \hline
\end{tabular}}
\caption{ROUGE scores for ClustSumm and Fusion summarization tasks, showing ClustSumm as one of the bottlenecks in E2ESumm performance.}
\label{tab:cluster_summ}
\end{table}
}

\newcommand{\OverallSumm}{
\begin{table}[t!]
\small
\resizebox{\columnwidth}{!}{\begin{tabular}{|l|c|}
\hline
    \textbf{Model}   & \textbf{ROUGE-1/2/L}   \\ \hline
BART-large \cite{lewis-etal-2020-bart}	& 28.17/8.61/24.01  \\ \hline
BART-large-aug 	&	\textbf{29.10/9.15/24.63}	\\ \hline
T5-base \cite{raffel2019exploring}	& 25.10/6.58/21.30	\\ \hline
\hline
BART-rel-oracle &	\textbf{30.98/10.61/26.22}	\\ \hline
\end{tabular}}
\caption{Model comparison for E2ESumm.}
\label{tab:overall_summ}
\end{table}
}

\newcommand{\RLResults}{
\begin{table}[t!]
\small
\resizebox{\columnwidth}{!}{\begin{tabular}{|c|c|c|c|}
\hline
    \textbf{Task}   & \textbf{ROUGE-1/2/L}  & \textbf{NLI}   & \textbf{Semantic Area}    \\ \hline
BART	&	 28.17/8.61/24.01  & 0.74  & 0.04 \\ \hline
BART-aug	&	29.10/9.15/24.63 & 0.77 & 0.01	\\ \hline
BART-aug + RL & 28.81/8.96/24.72 & 0.76 &  0.05	\\ \hline
\end{tabular}}
\caption{A comparison of model ROUGE, NLI, and Semantic Area scores.}
\label{tab:rl_results}
\end{table}
}

\newcommand{\ExampleSummariesb}{
\begin{table}[h!]
\small
\centering
\begin{tabularx}{\columnwidth}{|X|}
\hline
\textbf{Question:} I wonder how secure my deadbolt lock is.   How difficult is it for a professional to open such a lock?
 \\ \hline
\textbf{Answer 1:} Any lock can be opened. The questions are:    How long will it take? How much skill is required? What tools are needed? ...
\\ \hline
\textbf{Answer 2:}
In general, a professional is going to be able to open anything you have, because that's what they do all day. The reality though is that with the exception of high security locks like Medeco, it doesn't even take a professional to open them. ...
\\ \hline
\textbf{Answer 3:} Absolutely. Anyone with a bump key or lockpick can open a deadbolt. ... \\ \hline
\textbf{Answer 4:}  Bottom line is that if someone wants to get into your house .... they can. I've never seen a fool proof system. ... \\ \hline
\textbf{Answer ...:}  \\
\hline 
\hline
\textbf{BART Summary:} The answer to this question will be subjective and will depend on the type of lock. However, it is generally agreed that any lock can be opened by a professional and that it is not harder to pick than a normal lock.
 \\ \hline
\textbf{BART-aug Summary:} The answer to this question will depend on the type of lock and the tools needed. However, it is generally agreed that any lock can be opened by a professional.
 \\ \hline
\textbf{BART-aug+RL Summary:} Any lock can be opened. A deadbolt is more about resisting kicking open or using a credit card to slide in and raise the bolt. It's not so much about being harder to pick, as the lock mechanism in it is going to be very similar to a normal door handle.
 \\ \hline
\end{tabularx}
\caption{Additional sample input and model outputs. The outputs are factually consistent and cover multiple perspectives present in the input answers.}
\label{tab:example_summaries_2}
\end{table}
}

\newcommand*{\affaddr}[1]{#1}
\newcommand*{\affmark}[1][*]{\textsuperscript{#1}}

\title{AnswerSumm: A Manually-Curated Dataset and \\ Pipeline for Answer Summarization}

\author{
 \textbf{Alexander R. Fabbri}\affmark[$\dagger$]~\Thanks{Author is currently at Salesforce AI Research.}    \quad \textbf{Xiaojian Wu}\affmark[$\ddagger$]
 \quad \textbf{Srini Iyer}\affmark[$\ddagger$]
 \quad \textbf{Haoran Li}\affmark[$\ddagger$]
 \quad \textbf{Mona Diab}\affmark[$\ddagger$] \\
\affaddr{\affmark[$\dagger$] Yale University} 
  \affaddr{\affmark[$\ddagger$] Facebook AI} \\
  \texttt{afabbri@salesforce.com} \\
          \texttt{\{xiaojianwu,sviyer,aimeeli,mdiab\}@fb.com} 
}

\begin{document}
\maketitle
\begin{abstract}
Community Question Answering (CQA) fora such as Stack Overflow and Yahoo! Answers contain a rich resource of answers to a wide range of community-based questions. 
Each question thread can receive a large number of answers with different perspectives.
One goal of answer summarization is to produce a summary that reflects the range of answer perspectives. 
A major obstacle for this task is the absence of a dataset to provide supervision for producing such summaries. 
Recent works propose heuristics to create such data, but these are often noisy and do not cover all answer perspectives present. 
This work introduces a novel dataset of 4,631 CQA threads for answer summarization curated by professional linguists. 
Our pipeline gathers annotations for all subtasks of answer summarization, including relevant answer sentence selection, grouping these sentences based on perspectives, summarizing each perspective, and producing an overall summary.
We analyze and benchmark state-of-the-art models on these subtasks and introduce a novel unsupervised approach for multi-perspective data augmentation that boosts summarization performance according to automatic evaluation.
Finally, we propose reinforcement learning rewards to improve factual consistency and answer coverage and analyze areas for improvement.
\end{abstract}
\section{Introduction}\label{sec:introduction}
\IntroductionMotivatingExample
In a world of information overload and the ubiquity of discussion fora, there is a need for text summarization as a means of distilling relevant information into a concise form. 
The problem is even more pertinent for question answering within the context of Community Question Answering (CQA) fora, where a person poses a question and can get an abundance of answers to sift through. 
Ideally, an answer summary should cover the multiple perspectives found in the answers, where available. 
Table \ref{tab:example} illustrates such an example where a person poses a question about relocating to the US and obtaining a credit score and a credit card. 
We present a sample of the 8 answers to that question on StackExchange and a manually-curated summary covering the answers' main perspectives. 
Answer summarization is a form of query-based, multi-document summarization \cite{ernst2020superpal}, and creating answer summaries that reflect the underlying varying perspectives entails several subtasks: selection of answer sentences relevant to the question (query sentence relevance), grouping these sentences based on perspectives (clustering), summarizing each perspective (cluster summarization), and producing an overall fused summary (fusion). 
\par
To date, most CQA fora have a notion of a 'best answer,' which is either manually chosen by the person who asked the question or by a moderator, or obtained via community ratings.
Work in this field typically makes use of this best answer as a proxy for summaries, i.e. the focus is on extractive-like summaries \cite{tomasoni-huang-2010-metadata,chan-etal-2012-community,pande2013summarizing,wang-etal-2014-query, song2017summarizing}. 
Datasets such as WikiHowQA \citep{deng2020joint}, which consists of a question, a long answer, and an answer summary, focus on answer selection and the summarization of a single answer. 
While CQASumm \cite{chowdhury2019cqasumm} uses the chosen best answer as the answer summary, they also apply heuristics to ensure token overlap with the remaining answers.  However, the best answer only presents one person's perspective and rarely captures the variety of perspectives discussed in a thread. 
Furthermore, we find that the heuristics applied in CQASumm generally promote only long answers instead of multiple perspectives. 
To validate our hypothesis, we examine a set of 30 summaries from CQASumm and found that only 37\% of the examples contained multi-perspective answers. 
In contrast, 75\% of our dataset requires multi-perspective summaries. 
\par
As alluded to above, although answer summarization is an important research topic with practical applications, there are no relevant datasets or techniques to address it effectively, i.e. no manually-curated dataset exists for the answer summarization problem, and no dataset decomposes the task into its constituent subtasks. 
This work tries to close the research gap in answer summarization; we develop an annotation pipeline for multi-perspective abstractive answer summarization. 
We introduce the largest human-annotated dataset for answer summarization, containing components for sentence relevance, clustering, cluster summarization, and global answer summarization. 
We enlist ten professional linguists to contribute to our annotation efforts. 
We iterate over instructions and devise pre-pilot, pilot, and final annotation stages as well as re-annotation for quality assurance. 
We collect over 4,631 high-quality data points. 
For validation of our curated data set, we benchmark state-of-the-art models on the subtasks of this dataset and perform qualitative analysis to provide a clear baseline and directions for future work. 
We then propose a data augmentation pipeline to further boost summarization performance.
To generate a silver multi-perspective summarization dataset, we introduce a pipeline for automatically creating multi-perspective bullet-point answer summaries for data augmentation, which boosts performance.
We find that a strong baseline model trained on our human-annotated data inherently outputs factually consistent summaries, and model performance is improved by adding data from our automated pipeline. 
Finally, we introduce entailment-based and semantic area RL rewards namely to analyze its effect on factual consistency and semantic coverage, ensuring we are capturing \textit{all} factually relevant perspectives.
\footnote{For reproducibility of our findings, we will make our data and code publicly available at \url{https://github.com/Alex-Fabbri/AnswerSumm}.}
\section{Related Work}\label{sec:related_work}
\par \noindent
\textbf{Extractive Answer Summarization:}
Much work has focused on the extractive summarization setting as an answer-ranking problem \citep{chan-etal-2012-community,pande2013summarizing, wang-etal-2014-query}.
\citet{liu-etal-2008-understanding} find that only 48\% of the best answers on Yahoo! Answers are unique best answers; there are multiple correct ways to answer a question.
Other recent work has focused on sentence extraction using metadata \citep{tomasoni-huang-2010-metadata}, sparse-coding frameworks \cite{song2017summarizing}, or answer-aware sequential extraction \cite{hierarchical_sequential}. 
Our focus is on an answer summarization pipeline which ultimately results in abstractive answer summaries.
\par \noindent
\textbf{Abstractive Answer Summarization:}
Another line of work has attempted abstractive answer summarization by treating the tagged best answer as the gold summary of all the other answers \citep{chowdhury2019cqasumm, chowdhury2020neural}.
Recent work summarizes answers to medical questions via a medical concept graph \citet{zhang-etal-2020-summarizing} and incorporates multi-hop reasoning \cite{zhang-etal-2020-summarizing} and answer relevance from a QA model into the summarization model \cite{su-etal-2021-improve}. 
Most related to our dataset creation, \citet{chowdhury2019cqasumm} present CQASumm, a dataset of about 100k automatically-created examples consisting of the best answer as the gold summary, which, however, contains noise due to automatic creation. 
\par \noindent
\textbf{Multi-document Summarization:}
Answer summarization can be viewed as a query-based multi-document summarization (MDS) problem. 
Approaches to query-focused multi-document summarization have dealt with data sparsity via data augmentation \cite{pasunuru2021data} by restructuring the title and paragraphs of news articles to match the target task, coarse to fine-grained modeling \citet{xu-lapata-2020-coarse}, and by converting generic summarization data into proxy queries \cite{xu-lapata-2021-generating}
Several large-scale MDS datasets have been introduced in the news domain \citep{fabbri-etal-2019-multi, gu2020generating, ghalandari2020large}, for creating Wikipedia lead-paragraphs \citep{liu2019generating} and for long-form question answering \citep{fan-etal-2019-eli5}. 
However, Wikipedia summarization is topic-based and less granular than our setting, and the ELI5 dataset \citep{fan-etal-2019-eli5} summarizes web documents rather than direct query answers.

\section{AnswerSumm}
We introduce our annotation protocol and the characteristics of our manually-curated answer summarization dataset. 
Our annotation pipeline is illustrated in Figure \ref{fig:manual_pipeline}.
\ManualPipelineFigure
\paragraph{Annotation  Protocol}
Our annotation pipeline consists of four steps 1) Answer Sentence Selection (SentSelect), 2) Clustering (SentCluster), 3) Cluster Summarization (ClusterSumm), and 4) Cluster Summary Fusion (ClusterSummFusion).
We refer to the task of taking forum answers and producing final overall summaries \textbf{E2ESumm}. 
We believe that this pipeline mirrors the process by which humans create summaries of multiple answers by narrowing and organizing information, followed by paraphrasing.
Furthermore, dividing the summarization task in such a way paves the way for future work in understanding the steps by which a model creates a final summary, and recent work has similarly divided multi-document summarization into these subtasks \cite{ernst2020superpal}. 
For consistency, the same annotator completes all four steps for a given example.
However, we surmise that if each subtask is performed well, then multiple annotators can be involved for a given example.  
\par
For a given question thread, we present the annotator with the question, the forum from which the question came, the title of the post, and the tags that the original poster associated with the question. 
The user answers are then presented, where each answer has been automatically segmented into individual sentences using SpaCy \cite{spacy}. 
It is worth noting that sentence-level granularity is chosen as a simplifying assumption as an appropriate level of segmentation.
We are cognizant that clause level might be more accurate, however, given state-of-the-art clause detection as well as the precedence for sentence-level modeling in previous work \cite{tomasoni-huang-2010-metadata,song2017summarizing}, we opted for sentence-level segmentation.
\par
\textbf{Answer Sentence Selection (SentSelect):} We ask the annotators to mark each sentence as relevant or not depending on whether it provides information useful in answering the user's question. 
Annotators are instructed to mark as irrelevant sentences that do not function as independent units, such as those which need additional context to be understood as an answer to the question. 
As a result, noise from sentence segmentation may cause certain sentences to be marked as not relevant, but upon manual inspection, we found this to not be an issue. 
\par
\textbf{Clustering (ClusterSumm):} Annotators then cluster found relevant sentences into groups of the same topic. 
Sentences that are on the same topic but have different polarities are grouped together. 
We do not pre-define a desired number of clusters. 
Furthermore, clusters consisting of a single item are allowed, and a sentence can belong to multiple clusters. 
A sentence in multiple clusters may occur in the case of complex sentences which present multiple viewpoints. 
\par
\textbf{Cluster Summarization (ClustSumm):} The annotators summarize each individual cluster of relevant sentences from the previous step.
Each cluster summary should typically consist of 1-4 complete sentences.
To allow for abstract summaries, we instruct the annotators to try to use their own words (paraphrase) instead of copying large segments of the sentence clusters verbatim. Using the sentences' exact words is allowed, but they should not copy more than five consecutive words from a sentence. 
Additionally, the summary should function as an answer rather than as an analysis of the summary sentences. 
So, rather than stating, “Most of the answers indicate that it is highly subjective,” the annotator writes directly “It is highly subjective.” 
To ensure that the summary information can be found in the input answers, we also instruct the annotators to focus solely on the answer threads and not their external knowledge of the subject. 
The summary should solely (1) summarize the viewpoint present in the sentence cluster; and, (2) try to include some specific details from the assertions and anecdotes made by the answer sentences. 
We leave it to the annotator's judgment to leave out details from clusters that are too minute.
\par
\textbf{Cluster Summary Fusion (ClusterSummFusion):} The annotator combines the cluster summaries from the previous step into a single, coherent summary. 
The annotators can apply additional paraphrasing and need not simply insert each cluster summary; they may combine some cluster summaries into a single sentence. 
The annotator is asked to order and insert discourse connectives as necessary to increase inter-sentential coherence in the final summary. 
\paragraph{Data Filtering}
We selected question threads for annotation from the StackExchange data release\footnote{\url{https://archive.org/download/stackexchange}}, as it is publicly available and has been shared using a Creative Commons ShareAlike license. 
We created a whitelist of non-technical fora which do not require domain knowledge to summarize, similar to work on non-technical email summarization \citep{ulrich2008publicly}. 
We sampled from 38 fora. Table \ref{tab:num_sources} illustrates the top 20 fora and their frequency. 
In addition to this preliminary filtering, we further prompted annotators to discard any examples for which they felt unable to adequately assess the relevance of answer sentences to a question due to lack of required domain knowledge or context.
\par
The filtering of question threads was motivated by heuristics detailed in \citet{tomasoni-huang-2010-metadata}, which aims to find threads suitable for summarization. 
We only include answers with a non-negative community score which is determined by the number of upvotes by community members minus the number of downvotes. 
Moreover, they do not include comments to answers for simplicity, although future work may incorporate this into modeling. 
Threads were removed if 1) there were less than four answers, 2) the sum of the length of all answers was outside of (100, 1500) words, and 3) the average length of answers was outside of the (50, 300) words interval. 
Questions include the subject of the post and the content of the post when available.
Out of about 870k question threads, about 8k met these criteria. 
While this filtering may be strict, it avoids threads that contain short or single answers for which summarization may be superfluous, thus creating a higher-quality, diverse, dataset as confirmed by our analysis that 75\% of our examples require multi-perspective summaries . 
%
%
\paragraph{Quality Controls}
Our annotators are 10 professional linguists recruited through a professional vendor.
We provide the linguists with an example of an annotated question thread for clarity and discussed the instructions in-depth with the vendors to avoid ambiguities. 
To ensure that the linguists are well-trained and that the annotations meet our requirements, we completed our annotations in three stages. 
We began with a pre-pilot of 50 example question threads, followed by a pilot of 500 examples and then a final set of 5000 examples. 
We divide annotation files into groups of 50 examples, which are split among the annotators. 
We make use of the pilot and final annotation sets for our dataset release.
To determine inter-annotator agreement (IAA), 250 examples were repeated across three annotation files. 
A Fleiss Kappa of 0.25 was achieved for sentence relevance selection, the first task.
The IAA score indicates fair agreement. 
\paragraph{Dataset Statistics and Comparison}
 We provide statistics about the subtasks from our dataset pipeline in Table \ref{tab:statistics_answersumm}.
 \DatasetStatisticsAnswerSumm
 There does not exist a manually-curated dataset for abstractive answer summarization.
 CQASumm is the closest dataset with our desired answer summarization qualities, although it is created automatically based on heuristics which simply promote answers as summaries rather than truly summarizing answers. 
 We also present a comparison of dataset statistics between our dataset AnswerSumm, and the standard XSum and CNN-Daily Mail \cite{nallapati-etal-2016-abstractive} summarization datasets in Table \ref{tab:statistics}.
 In general, we find our dataset to be more abstractive than CNN-DailyMail and less so than XSum.
Furthermore, the average number of input tokens for the E2ESumm task, is larger than those two datasets, confirming that the input to our tasks provides reasonable grounds for requiring summarization. 
\ForumCounts
\DatasetComparison
\section{Pipeline for Data Augmentation}\label{sec:dataset_pipeline}
Manually annotating data at the scale of other existing summarization datasets such as CNN-DailyMail is impractical. 
Taking advantage of the abundance of unlabeled StackExchange fora available, we develop a pipeline to automatically create data similar to that which is manually annotated above. 
This process provides augmented data for training summarization models.
\paragraph{Data Filtering}
Similar to filtering for manual annotation, we obtained question threads from StackExchange and applied heuristics motivated by \citet{tomasoni-huang-2010-metadata} to find threads suitable for summarization. 
Threads are removed if: 1) there are less than three answers; 2) the longest answer is at least 400 words; 3) the input token length of all answers is not between 100 and 1000 words; and, 4) the average length of answers is between 50 and 300 words. Heuristics were chosen to provide enough examples for data augmentation, leaving about 130k question threads in total. 
\PipelineFigure
\paragraph{Pipeline Overview}
The input to our pipeline is a user question and its answers. 
We select question threads from StackExchange and operate on the sentence-level of these answers, as in our manually-created data. 
Our automatic dataset pipeline consists of the following components which aim to mirror the manual pipeline: 1) a relevance model to select relevant sentences and remove irrelevant ones; 2) a clustering model to cluster similar content -- reflecting various perspectives; and, 3) input and abstractive summary creation from cluster centroids, resulting in bullet points for the various perspectives reflected in the answers.
Figure \ref{fig:pipeline} illustrates the pipeline.
\paragraph{Relevance model:}
A sentence-level relevance model trained on CQA fora is leveraged to eliminate irrelevant sentences from the input (collection of answers to a question). The output from this stage serves as input to the clustering stage. Model details are found in Section \ref{sec:results}.
\paragraph{Clustering:}
Typical K-Means clustering for short text \citep{xu2017self,hadifar-etal-2019-self,rakib2020enhancement} does not work for our setting as the value of K is not known a priori. 
In fact, it varies from question to question. 
Accordingly, we use the sentence-transformers library \citep{reimers-2019-sentence-bert} to perform clustering.
Specifically, we start with a RoBERTa-based model fine-tuned for sentence embeddings on an entailment dataset, which is further fine-tuned for semantic similarity. 
Clustering parameters are chosen based on a StackOverflow clustering dataset containing labeled clusters, as provided in \citet{rakib2020enhancement}. 
We apply Agglomerative clustering with average linkage, cosine distance, and a maximum distance of $0.65$. Parameters are empirically chosen.
\par
To create the final summaries, we locate the centroid of clusters with at least two sentences and select these centroids as bullet-point summaries. 
Further, we remove the centroid sentences from the sentence-segmented input answers to create a challenging abstractive summarization dataset analogous to the XSum dataset \citep{narayan-etal-2018-dont}. 
Since each cluster contains at least two sentences, we assume that given a perfect clustering algorithm, a related sentence can help generate the removed centroid sentence. 
While removing sentences naturally decreases coherence, we believe that this introduces a tolerable level of noise. 
We also experimented with cluster centroid paraphrasing and not removing from the input, but this did not improve downstream performance, which we use to measure the value of this dataset and the level of noise.
\section{RL-Based Training}\label{sec:dataset_pipeline}
Cross-entropy loss in standard sequence-to-sequence model training suffers from exposure bias and also does not directly optimize evaluation metrics \cite{ranzato2015sequence}.
The REINFORCE algorithm \cite{Williams:92}, on the other hand, allows for optimizing the evaluation metrics using non-differentiable rewards. 
We use an RL multi-reward objective to promote summaries with both high coverage of the input answers and faithfulness. 
\subsection{Multi-Reward Optimization}
We follow the settings of \citet{pasunuru-bansal-2018-multi} for optimizing multiple rewards. 
In the equations which follow, $x = \{x_1,\: x_2,\: \dots,\: x_{n'}\}$ refers to the input source tokens (e.g. a question and its answers), and $y^{*} = \{y^{*}_1,\: y^{*}_2,\: \dots,\: y^{*}_{N}\}$ refers to the gold target summary which consists of $\{y^{*}_{1_s},\: y^{*}_{s_s},\: \dots,\: y^{*}_{N_s}\}$ sentences. 
Standard training minimizes the negative log-likelihood (NLL) loss using teacher forcing \cite{williams1989learning}:
\begin{equation}
      L_{ml} =  -\sum_{t=1}^{N} \log p(y^{*}_{t}|y^{*}_1,...,y^{*}_{t-1}, x)
\end{equation}
For our RL optimization, we use self-critical policy gradient training as in \citet{paulus2017deep, rennie2017self}. 
At each time-step, we produce an output $y^{s}$ by sampling from the current decoding probability, $p(y^{s}_{t}|y^{s}_1,...,y^{s}_{t-1}, x)$, as well as an output $\hat{y}$ obtained by greedily decoding from the current probability distribution. 
We define a reward function $r(y, x, y^{*}) \in [0,1]$, i.e., the reward function compares $y$ with $x$ and $y^{*}$. The RL loss function $L_{rl}(x,y^*)=$: 

\begin{equation}
      \resizebox{\columnwidth}{!}{$(r(\hat{y}, x, y^*) - r(y^{s}, x, y^*))  \sum_{t=1}^{N} \log p(y^{s}_{t}|y^{s}_1,...,y^{s}_{t-1}, x)$}
\end{equation}
As in \citet{paulus2017deep} and \citet{pasunuru-bansal-2018-multi}, we use a mixture of the two losses above:
\begin{equation}
     L_{mixed}  = \gamma_{rl} L_{rl} + \gamma_{ml} L_{ml}, 
\end{equation}
where $ \gamma_{rl}$ and $\gamma_{ml}$ are tunable hyperparameters used as scaling factors. 
Rather than applying weights to each reward, we follow \citet{pasunuru-bansal-2018-multi} and optimize $L_{mixed}$ by alternating rewards in each minibatch. 
\subsection{Rewards}
We use the following RL reward functions: (1) textual entailment (NLI) for faithfulness, and (2) semantic area to measure the coverage of a summary in a semantic space. 
\paragraph{NLI for Faithful Summarization:} We use the degree of entailment of summaries given input answers as a reward to promote faithfulness of answer summarization.
\citet{falke-etal-2019-ranking} define NLI as a measure of faithfulness for ranking summaries as follows: Let $\mathcal{N}$ be an NLI model which, given a claim $c$ and a premise $p$, computes $\mathcal{N}(p, c)$, the probability that the claim is entailed by the premise. 
We use this to calculate the NLI score for a summary $y$ consisting of $N_s$ sentences:
\begin{equation}
      \text{NLI}(y, x) =  \frac{1}{N_{s}} \sum_{i=1}^{N_{s}}  \max_{s \in x} \mathcal{N}(s, y_{i_s}) 
\label{eq:nli}
\end{equation}
\paragraph{Semantic Area for Multi-Perspective Summarization:} We aim to reward summaries that include more of the perspectives found in the input answers. 
To achieve diverse extractive summarization, \citet{yogatama-etal-2015-extractive} embed sentences in the semantic space and then select those sentences whose convex hull maximizes the volume in that space.
This idea of semantic volume is also used to measure the semantic overlap between summaries and references in \citet{jung-etal-2019-earlier}. 
We use semantic volume as a proxy for covering multiple perspectives; the summary with the larger semantic volume covers a wider range of views discussed in the input. 
We make use of sentence-transformers \cite{reimers-gurevych-2019-sentence} to obtain sentence embeddings for each sentence.
We project each embedding onto two dimensions using Principal Component Analysis (PCA) as in \citet{jung-etal-2019-earlier}, and thus, our volume calculation reduces to an area calculation, which we refer to as \textbf{Semantic Area}. 
We use min-max normalization to keep the reward between 0 and 1. 
We split the dataset into training, validation, and testing sets of size 3131, 500, and 1000 examples. 
For relevance labeling, we train RoBERTa  \cite{liu2019roberta} for binary relevance classification with the user question and sentence as inputs. 
We train with a polynomial decay learning rate scheduler with learning rate $2\mathrm{e}{-5}$, using the Adam optimizer \citep{kingma2014adam} for three epochs. 
We compare this model to one trained on the ANTIQUE \citep{hashemi2020antique} relevance data for query-sentence relevance. 
The data consists of Yahoo! answers and relevance labels on a scale from 1-4, with 1-2 not relevant and 3-4 relevant. 
\par
For experiments in ClusterSumm and E2ESumm, our baseline abstractive text summarization model is BART \cite{lewis-etal-2020-bart}, a pretrained denoising autoencoder that builds off of the sequence-to-sequence transformer of \citet{vaswani2017attention}. 
For E2ESumm results, our primary focus, we also apply several state-of-the-art abstractive summarization models such as T5-base \cite{raffel2019exploring}. 
For the cluster summarization task, the input is the individual sentences clustered by the annotators, while for the cluster fusion step, the input is the concatenation of the cluster summaries. 
For E2ESumm, input to the models is the question concatenated with input answers. 
For both summarization tasks, we fine-tune BART using a polynomial decay learning rate scheduler with learning rate \num{3e-5}, using the Adam optimizer \citep{kingma2014adam}. 
We train with 500 warm-up steps and 20,000 total steps and pick the model with the best label-smoothed cross-entropy \cite{szegedy2016rethinking} validation loss. 
T5 is trained for 3 epochs with a linear learning rate scheduler. 
In RL experiments, we train using BART from scratch, as opposed to using a model already fine-tuned on answer summarization, as we found that this model better learned to follow the given rewards. 
Following similar ratios in \citet{lu2019multi}, we set ($\gamma_{rl}$,$\gamma_{ml}$) =  (0.9, 0.1).
Hyperparameters are tuned on the validation set.
\section{Results \& Discussion}\label{sec:results}
We provide strong baseline results for the SentSelect, ClusterSumm, ClusterSummFusion, and E2ESumm subtasks of AnswerSumm as a basis for future work. 
\par
\RelResults
The best results for SentSelect are yielded by RoBERTa relevance classification as illustrated in Table \ref{tab:rel_results}.
RoBERTa yields an F1 score of 0.49. 
Despite being the highest, the relatively low result points to the difficulty and subjectivity of selecting relevant sentences for community question answering fora.
This is further supported by the observed low IAA of fair agreement (Fleiss Kappa of 0.25). 
Moreover, concatenating the sentences labeled as relevant on the test set as a final summary results in long summaries with high recall (82.81 ROUGE-1 Recall). 
This suggests that much of the important information to be summarized can be captured by this relevance model. 
The ANTIQUE-trained model obtains an F1 score of 0.41 and notably predicts many false positives (71\%). 
While this model performs worse on this relevance classification task, we find that using this trained model for automatically generated data allows for an improved downstream summarization model, when compared to the better classifier trained solely on our manually-annotated data. 
Accordingly, we opt for using the ANTIQUE-trained model in our overall summarization task. 
The improved performance is likely due to more sentences being labeled as relevant (implicitly encoding a recall bias), allowing for more sentences to be sent to the clustering algorithm, a noisy step itself, ensuring better quality clusters. 
\par
Results for ClusterSumm and ClusterSummFusion are shown in Table \ref{tab:cluster_summ}. 
These results point to the difficulty of ClusterSumm as one of the sources of difficulty for E2ESumm performance, as ClusterSummFusion can be done fairly easily. 
We believe that some difficulties found in ClusterSumm are also found in the E2ESumm task.
\ClusterSumm
\OverallSumm
\par
The results for E2ESumm are presented in Table \ref{tab:overall_summ}. 
BART-large outperforms T5 model, but scores are rather low when compared to the extractive oracle above. 
To investigate this further, we train a BART-only model using the question concatenated with the oracle relevant sentences chosen by the annotators, BART-rel-oracle.
BART-rel-oracle significantly outperforms the vanilla model. 
This suggests that improved content selection would boost performance. 
However, we believe that the primary cause of the low performance is the difficulty in learning the compression rate and abstractiveness of the gold summaries. 
The percentage of novel uni-grams in BART is only 4\%, as opposed to the 21\% present in the gold summaries. 
This suggests that despite being trained on more abstractive data, BART is not learning (not generalizing) how to abstract well enough. 
We also note the model trained on additional augmented data through our automatic pipeline, BART-aug, achieves a large performance boost compared to vanilla BART, thereby validating the efficacy of our automatic pipeline for potential applications to new domains. 
It should be noted that we experimented with augmenting our manually-curated data with data from CQASumm, but performance did not improve over vanilla BART. 
Hence, the task is indeed sensitive to the quality and type of data used for augmentation.
\par
The results of adding RL rewards to BART trained with data augmentation are shown in Table \ref{tab:rl_results}. 
Both BART with augmented data and RL rewards achieve higher NLI scores than the baseline, while only the model with RL rewards obtains a higher Semantic Area score. 
The improved ROUGE score for BART-aug likely results from training on additional data that resembles the target domain, as in \citet{fabbri2020improving}, while noise in the unsupervised data may reduce the Semantic Area scores.
The addition of RL rewards improves the semantic area score over the augmented model, although the slight decrease in ROUGE-1/2 show that semantic area does not completely align with ROUGE score. 
We analyzed 25 model outputs for factual consistency and found that the models are largely factually consistent, and very extractive (BART-aug + RL having the fewest novel unigrams at 3.9\$). 
This suggests these differences in NLI score do not exhibit a large qualitative difference in faithfulness, and the lower NLI score of the RL model may be from the introduction of the semantic area reward. 
Also, we note that the gold summaries themselves have low NLI and Semantic Area scores of 0.46 and 0.03. 
As the gold summaries are more abstractive, the entailment relationship between them and the input may not be as straightforward as the primarily extractive model outputs.
This phenomenon suggests the need for improved metrics and rewards for abstractive factual consistency and semantic coverage. 
We provide example summaries in the supplementary materials.

\RLResults
\section{Conclusion and Future Work}\label{sec:conclusion}
We develop an annotation pipeline for multi-perspective answer summarization, introducing the largest human-annotated dataset for this task. 
We benchmark state-of-the-art models on the content selection, cluster summarization, and end-to-end summarization subtasks of this dataset. 
We also introduce a pipeline for answer summarization data augmentation that boosts summarization performance. 
Through an analysis of the effects of reinforcement-learning rewards and qualitative examination of model outputs, we point to difficulties in these tasks and areas for future improvement in content selection, abstraction levels, and metrics for model comparison. 
\section{Ethical Considerations}\label{sec:ethical_considerations}
As we propose a novel conversation summarization dataset creation pipeline and modeling components, this section is divided into the following two parts.
\subsection{New Dataset}
\paragraph{Intellectual Properties and Privacy Rights}
We make use of publicly-available StackExchange data for all our annotations. 
We manually reviewed our dataset output for quality and potential problems. 
\paragraph{Compensation for Annotators}
Compensation was determined by standard in-house rates, amounting to about \$6 per data point collected. 
\subsection{NLP Application}
\paragraph{Bias}
Biases may exist in the datasets, such as political bias and gender bias in Yahoo! Answers. 
Thus, models trained on these datasets may propagate these biases.
\paragraph{Misuse Potential and Failure Mode}
When used as intended, applying the summarization models described in this paper can save people much time. 
However, the current models are still prone to producing hallucinated summaries, and in such a case, they may contribute to misinformation on the internet. 
We move the needle in faithful summarization in this paper, but further research is needed to ensure the faithfulness of abstractive summaries to address this issue, as this issue is present among all current abstractive summarization models. 
\paragraph{Environmental Cost}
The experiments described in the paper make use of V100 GPUs. We used up to 8 GPUs per experiment. 
The experiments may take several hours. 
Several dozen experiments were run due to parameter search, and future work should experiment with distilled models for more light-weight training. 
We note that while our work required extensive experiments to draw sound conclusions, future work will be able to draw on these insights and need not run as many large-scale comparisons. 
Models in production may be trained once for use using the most promising settings.  
\bibliography{arr}
\bibliographystyle{acl_natbib}

\appendix
\section{Appendix}\label{sec:appendix}
We provide sample model outputs in Tables \ref{tab:example_summaries_1} and \ref{tab:example_summaries_2} which characterize the factual consistency and multi-perspective nature of the models.
\ExampleSummaries
\ExampleSummariesb

\end{document}


\maketitle

\section{Appendix}\label{sec:appendix}
We provide sample model outputs in Tables \ref{tab:example_summaries_1} and \ref{tab:example_summaries_2} which characterize the factual consistency and multi-perspective nature of the models.
\ExampleSummaries
\ExampleSummariesb
